\documentclass{llncs}
\usepackage{url}
\usepackage{booktabs}
\usepackage{graphicx}
\usepackage{float}
\usepackage{enumitem}

\begin{document}

\title{In Search of Credible News}
\titlerunning{In Search of Credible News}
\author{Momchil Hardalov\inst{1}
\and Ivan Koychev\inst{1}
\and Preslav Nakov\inst{2}
}
\authorrunning{Momchil Hardalov et al.} % abbreviated author list (for running head)
%
%%%% list of authors for the TOC (use if author list has to be modified)
\tocauthor{Momchil Hardalov, Ivan Koychev, Preslav Nakov}
\institute{FMI, Sofia University ``St. Kliment Ohridski'', Sofia, Bulgaria,\\
\email{momchil.hardalov@gmail.com,koychev@fmi.uni-sofia.bg}
\vspace{2pt}
\and
Qatar Computing Research Institute, HBKU, Doha, Qatar,\\
\email{pnakov@qf.org.qa}}

\maketitle              % typeset the title of the contribution

\begin{abstract}
We study the problem of finding fake online news. This is an important problem as news of questionable credibility have recently been proliferating in social media at an alarming scale.
As this is an understudied problem, especially for languages other than English, we first collect and release to the research community three new balanced credible vs. fake news datasets derived from four online sources.
We then propose a language-independent approach for automatically distinguishing credible from fake news, based on a rich feature set. In particular, we use linguistic ($n$-gram), credibility-related (capitalization, punctuation, pronoun use, sentiment polarity), and semantic (embeddings and DBPedia data) features.
Our experiments on three different testsets show that our model can distinguish credible from fake news with very high accuracy.

\end{abstract}

\section{Introduction}

Internet and the proliferation of smart mobile devices have changed the way information spreads, e.g.,
social media, blogs, and micro-blogging services such as Twitter,
Facebook
and Google+
have become some of the main sources of information for millions of users on a daily basis. 
On the positive side, this has democratized and accelerated content creation and sharing.
On the negative side, it has made people vulnerable to manipulation, as the information in social media is typically not monitored or moderated in any way.
Thus, it has become increasingly harder to distinguish real news from misinformation, rumors, unverified, manipulative, and even fake content. Not only are online blogs nowadays flooded by biased comments and fake content, but also online news media in turn are filled with unreliable and unverified content, e.g., due to 
the willingness of journalists to be the first to write about a hot topic, often by-passing the verification of their information sources; there are also some online information sources created with the sole purpose of spreading manipulative and biased information. 
Finally, the problem extends beyond the cyberspace, as in some cases, fake news from online sources have crept into mainstream media.

Journalists, regular online users, and researchers are well aware of the issue, and topics such as information credibility, veracity, and fact checking are becoming increasingly important research directions \cite{journals/intr/CastilloMP13,phdthesis:graves,Zubiaga2014}. 
For example, there was a recent 2016 special issue of the ACM Transactions on Information Systems journal on Trust and Veracity of Information in Social Media \cite{Papadopoulos:2016:OSI}, and there is an upcoming SemEval-2017 shared task on rumor detection.

As English is the primary language of the Web, most research on information credibility and veracity has focused on English, while other languages have been largely neglected. To bridge this gap, below we present experiments in distinguishing real from fake news in Bulgarian; yet, our approach is in principle language-independent. In particular, we distinguish between real news vs. fake news that in some cases are designed to sound funny (while still resembling real ones); thus, our task can be also seen as humor detection \cite{mihalcea-strapparava:2005:HLTEMNLP,yang-EtAl:2015:EMNLP2}.

As there was no publicly available dataset that we could use, we had to create one ourselves. We collected two types of news: credible, coming from trusted online sources, and fake news, written with the intention to amuse, or sometimes confuse, the reader who is not knowledgeable enough about the subject. We then built a model to distinguish between the two, which achieved very high accuracy.

The remainder of this paper is organized as follows: Section~\ref{sec:relatedwork} presents some related work.
Section~\ref{sec:method} introduces our method for distinguishing credible from fake news. Section~\ref{sec:experandeval} presents our data, feature selection, the experiments, and the results. Finally, Section~\ref{sec:futurework} concludes and suggests directions for future work.

\section{Related Work}
\label{sec:relatedwork}

Information credibility in social media is studied by Castillo \& al. \cite{journals/intr/CastilloMP13}, who formulate it as a problem of finding false information about a newsworthy event. They focus on tweets using variety of features including user reputation, author writing style, and various time-based features.

Zubiaga \& al. \cite{zubiaga2015analysing} studied how people handle rumors in social media. They found that users with higher reputation are more trusted, and thus can spread rumors among other users without raising suspicions about the credibility of the news or of its source.

Online personal blogs are another popular way to spread information by presenting personal opinions, even though researchers disagree about how much people trust such blogs. Johnson \& al. \cite{johnson2007every} studied how blog users act in the time of newsworthy event, e.g., such as the crisis in Iraq, and how biased users try to influence other people.

It is not only social media that can spread information of questionable quality. The credibility of the information published on online news portals has also been questioned by a number of researchers
\cite{brill2001online,ketterer1998teaching,finberg2002digital}. As timing is a crucial factor when it comes to publishing breaking news, it is simply not possible to double-check the facts and the sources, as is usually standard in respectable printed newspapers and magazines. This is one of the biggest concerns about online news media that journalists have \cite{cassidy2007online}.

The interested reader can see \cite{Zaharia:2010:SCC:1863103.1863113} for a review of various methods for detecting fake news, where different approaches are compared based on linguistic analysis, discourse, linked data, and social network features.

Finally, we should also mention work on humor detection. Yang \& al. \cite{yang-EtAl:2015:EMNLP2} identify semantic structures behind humor, and then design sets of features for each structure; they further develop anchors that enable humor in a sentence. However, they mix different genres such as news, community question answers, and proverbs, as well as the One-Liner dataset \cite{mihalcea-strapparava:2005:HLTEMNLP}. In contrast, we focus on news both for positive and for negative examples, and we do not assume that the reason for a news being not credible is the humor it contains.

\section{Method}
\label{sec:method}

We propose a language-independent approach for automatically distinguishing credible from fake news, based on a rich feature set.
In particular, we use \emph{linguistic} ($n$-gram), \emph{credibility} (capitalization, punctuation, pronoun use, sentiment polarity), and \emph{semantic} (embeddings and DBPedia data) features.

\subsection{Features}
\label{subsec:features}

\subsubsection{\bf Linguistic ($n$-gram) Features}

Before generating these features, we first perform initial pre-processing: tokenization and stop word removal. We define stop words as the most common, functional words in a language (e.g., conjunctions, prepositions, interjections, etc.); while they fit well for problems such as author profiling, they turn out not to be particularly useful for distinguishing credible from fake news.
Eventually, we experimented with the following linguistic features:
\begin{itemize}
\item \textbf{$n$-grams:} presence of individual uni-grams and bi-grams. The rationale is that some $n$-grams are more typical of credible vs. fake news, and vice versa;
\item \textbf{tf-idf:} the same $n$-grams, but weighted using tf-idf;
\item \textbf{vocabulary richness:} the number of unique word types used in the article, possibly normalized by the number of word tokens.
\end{itemize}

\subsubsection{\bf Credibility Features}

We also used the following credibility features, which were previously proposed in the literature \cite{journals/intr/CastilloMP13}:

\begin{enumerate}
	\item Length of the article (number of tokens); \label{itm:wordsCount}
    \item Fraction of words that only contain uppercase letters; \label{itm:allUpper}
    \item Fraction of words that start with an uppercase letter; \label{itm:startUpper}
    \item Fraction of words that contain at least one uppercase letter; \label{itm:hasUpper}
    \item Fraction of words that only contain lowercase letters; \label{itm:lowerCase}
    \item Fraction of plural pronouns; \label{itm:pluralPronouns}
    \item Fraction of singular pronouns; \label{itm:singualrPronouns}
    \item Fraction of first person pronouns; \label{itm:firstPerson}
    \item Fraction of second person pronouns; \label{itm:secondPerson}
    \item Fraction of third person pronouns; \label{itm:thirdPerson}
    \item Number of URLs; \label{itm:urls}
    \item Number of occurrences of an exclamation mark; \label{itm:exclMarks}
    \item Number of occurrences of a question mark; \label{itm:questionMarks}
    \item Number of occurrences of a hashtag; \label{itm:hashtag}
    \item Number of occurrences of a single quote; \label{itm:singleQuote}
    \item Number of occurrences of a double quote. \label{itm:doubleQuote}
\end{enumerate}

We further added some sentiment-polarity features from lexicons generated from Bulgarian movie reviews \cite{kapukaranov2015fine} (5,016 positive, and 2,415 negative words), which we further expanded with some more words. Based on these lexicons, we calculated the following features:
\begin{enumerate}
\setcounter{enumi}{16 	}
    \item Proportion of positive words; \label{itm:positiveWords}
    \item Proportion of negative words; \label{itm:negativeWords}
    \item Sum of the sentiment scores for the positive words; \label{itm:scorePositive}
    \item Sum of the sentiment scores for the negative words. \label{itm:scoreNegative}
\end{enumerate}

Note that we eventually ended up using only a subset of the above features, as we performed feature selection as described in Section~\ref{subsec:featselect} below.

\subsubsection{\bf Semantic (Embedding and DBPedia) Features} 

Finally, we use embedding vectors to model the semantics of the documents. We wanted to model implicitly some general world knowledge, and thus we trained word2vec vectors on the text of the long abstracts from the Bulgarian DBPedia.\footnote{\url{http://wiki.dbpedia.org/}} Then, we built vectors for a document as the average of the word2vec vectors of the non-stop word tokens it is composed of.

\subsection{Classification}

As we have a rich set of partially overlapping features, we used logistic regression for classification with L-BFGS \cite{liu1989limited} optimizer and elastic net regularization \cite{zou2005regularization}, which combines L1 and L2 regularization. This classification setup converges very fast, fits well in huge feature spaces, is robust to over-fitting, and handles overlapping features well.
We fine-tuned the hyper-parameters of our classifier (maximum number of iterations, elastic net parameters, and regularization parameters) on the training dataset. We further applied feature selection as described below.

\section{Experiments and Evaluation}
\label{sec:experandeval}

\subsection{Data}
\label{subsec:data}

As there was no pre-existing suitable dataset for Bulgarian, we had to create one of our own. For this purpose, we collected a diverse dataset with enough samples in each category. We further wanted to make sure that our dataset will be good for modeling credible vs. fake news, i.e., that will not degenerate into related tasks such as topic detection (which might happen if the credible and the fake news are about different topics), authorship attribution (which could be the case if the fake news are written by just 1-2 authors) or source prediction (which can occur if all credible/fake news come from just one source).
Thus, we used four Bulgarian news sources (from which we generated one training and  three separate balanced testing datasets):
\begin{enumerate}
\item We retrieved most of our credible news from \textbf{Dnevnik},\footnote{\url{http://www.dnevnik.bg/}} a respected Bulgarian newspaper; we focused mostly on politics. This dataset was previously used in research on finding opinion manipulation trolls \cite{mihaylov-georgiev-nakov:2015:CoNLL,mihaylov-EtAl:2015:RANLP2015,ACL2016:trolls}, but its news content fits well for our task too (\textit{\textbf{5,896} credible news});
\item As our main online source of fake news, we used a website with funny news called \textbf{Ne!Novinite}.\footnote{\url{http://www.nenovinite.com/}} We crawled topics such as politics, sports, culture, world news, horoscopes, interviews, and user-written articles (\textit{\textbf{6,382} fake news});
\item As an additional source of fake news, we used articles from the \textbf{Bazikileaks}\footnote{\url{https://neverojatno.wordpress.com/}} blog. These documents are written in the form of blog-posts and the content may be classified as ``fictitious'', which is another subcategory of fake news. The domain is politics (\textit{\textbf{656} fake news});
\item And finally, we retrieved news from the \textbf{bTV  Lifestyle section},\footnote{\url{http://www.btv.bg/lifestyle/all/}} which contains both credible (in the \emph{bTV} subsection) and fake news (in the \emph{bTV Duplex} subsection). In both subsections, the articles are about popular people and events (\textit{\textbf{69} credible  and \textbf{68} fake news});
\end{enumerate}

We used the documents from Dnevnik and Ne!Novinite for training and testing: 70\% for training and 30\% for testing. We further had two additional test sets: one of bTV vs. bTV Duplex, and one on Dnevnik vs. Bazikileaks. All test datasets are near-perfectly balanced. 

Finally, as we have already mentioned above, we used the long abstracts in the Bulgarian DbPedia to train word2vec vectors, which we then used to build document vectors, which we used as features for classification. (\textit{\textbf{171,444} credible samples}).

\subsection{Feature Selection}
\label{subsec:featselect}

\begin{figure}[H]
  \includegraphics[width=\textwidth,  height=8cm]{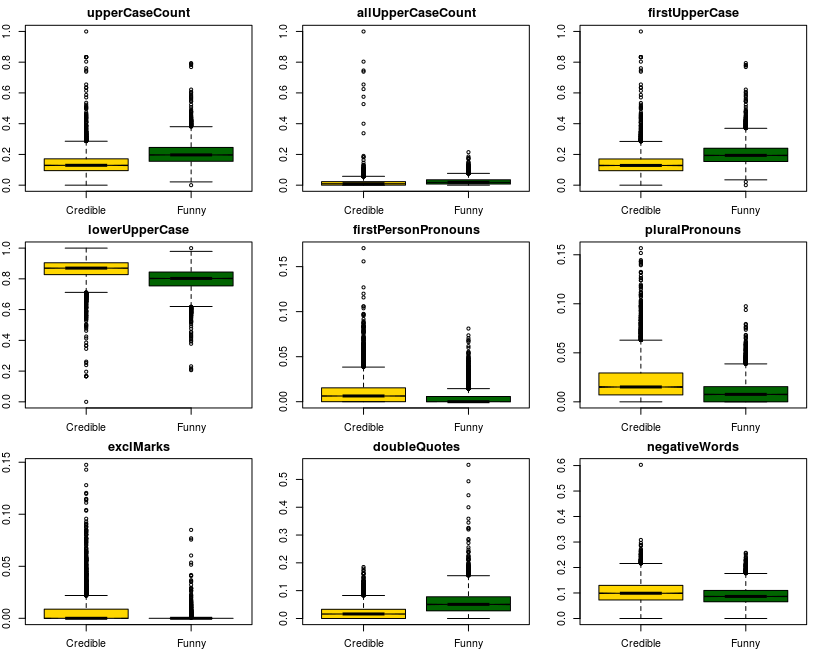}
  \caption{Boxplots presenting the distributions of some credibility features in credible vs. funny news.}
  \label{fig:featureSummary}
\end{figure}

We performed feature selection on the credibility features. For this purpose, we first used Learning Vector Quantization (LVQ) \cite{kohonen1990improved} to obtain a ranking of the features from Section~\ref{subsec:features} by their importance on the training dataset; the results are shown in Table~\ref{tab:featureSelection}.
See also Figure~\ref{fig:featureSummary} for a comparison of the distribution of some of the credibility features in credible. vs. funny news.

\begin {table}[H]
  \begin{center}
    \begin{tabular}{ p{4cm}p{3cm} }
      \hline
      Features & Importance \\
      \hline
      doubleQuotes \ref{itm:doubleQuote} & 0.7911 \\ 
      upperCaseCount \ref{itm:hasUpper} & 0.7748 \\ 
      lowerUpperCase \ref{itm:lowerCase} & 0.7717 \\ 
      firstUpperCase \ref{itm:startUpper} & 0.7708 \\ 
      pluralPronouns \ref{itm:pluralPronouns} & 0.6558 \\ 
      firstPersonPronouns \ref{itm:firstPerson} & 0.6346 \\ 
      allUpperCaseCount \ref{itm:allUpper} & 0.6282 \\ 
      negativeWords \ref{itm:negativeWords} & 0.5944 \\ 
      positiveWords \ref{itm:positiveWords} & 0.5834 \\ 
      tokensCount \ref{itm:wordsCount} & 0.5779 \\ 
      singularPronouns \ref{itm:singualrPronouns} & 0.5286 \\ 
      thirdPersonPronouns \ref{itm:thirdPerson} & 0.5273 \\ 
      negativeWordsScore \ref{itm:scoreNegative} & 0.5206 \\ 
      hashtags \ref{itm:hashtag} & 0.4998 \\ 
      urls \ref{itm:urls} & 0.4987\\ 
      positiveWordsScore \ref{itm:positiveWords} & 0.4910 \\ 
      singleQuotes \ref{itm:singleQuote} & 0.4884 \\ 
      secondPersonPronouns \ref{itm:secondPerson} & 0.4408 \\ 
      questionMarks \ref{itm:questionMarks} & 0.4407 \\ 
      exclMarks \ref{itm:exclMarks} & 0.3160 \\
      \bottomrule[1.25pt]
    \end{tabular}
    \caption {Features ranked by the LVQ importance metric.}
    \label{tab:featureSelection}
  \end{center}
\end {table}

Then, we experimented with various feature combinations of the top-ranked features, and we selected the combination that worked best on cross-validation on the training dataset (compare to Table~\ref{tab:featureSelection}):

\begin{itemize}
\item Fraction of negative words in the text (negativeWords);
\item Fraction of words that contain uppercase letters only (allUpperCaseCount);
\item Fraction of words that start with an uppercase letter (firstUpperCase);
\item Fraction of words that only contain lowercase letters (lowerUpperCase);
\item Fraction of plural pronouns in the text (pluralPronouns);
\item Number of occurrences of exclamation marks (exclMarks);
\item Number of occurrences of double quotes (doubleQuotes).
\end{itemize}

\begin {table}[H]
  \begin{center}
    \begin{tabular}{ p{6cm}p{2cm}p{2cm}p{2cm} }
      \hline
	  Feature Groups & Dnevnik & bTV vs.  & Dnevnik vs.  \\
      & Ne!Novinite & bTV Duplex & Bazikileaks \\
	  \hline
      
      \textbf{Credibility + Linguistic + Semantic} & \textbf{99.36} & 62.04 & \textbf{85.53} \\

      \textbf{Credibility + Semantic}  & 92.67 & \textbf{75.91} & 82.99 \\

      Linguistic + Credibility & 96.02 & 59.12 & \textit{61.94} \\

       \textbf{Semantic} & 98.95 & 61.31 & 71.01 \\

      Linguistic & 95.71 & \textit{56.93} & 73.25 \\

      Credibility & \textit{83.25} & 62.04 & 79.85 \\

      \hline
      Baseline (majority class) &  52.60 & 50.36  & 50.86\\
      \bottomrule[1.25pt]
    \end{tabular}
    \caption {Accuracy for different feature group combinations.}
    \label{tab:evaluationResults}
  \end{center}
\end {table}

\subsection{Results}

Table \ref{tab:evaluationResults} shows the results when using all feature groups and when turning off some of them. We can see that the best results are achieved when experimenting with ``Credibility + Semantic'' and ``Credibility + Linguistic + Semantic'' feature combinations, and the results are worse when only using credibility and linguistic features. 

Analyzing the results on the Dnevnik vs. the Ne!Novinite testset (first column), we can see that the linguistic features are more important than the credibility ones. Yet, the semantic features are even more important. When we combine all the feature groups, we achieve 99.36\% accuracy, but this is only marginally better than using the semantic features alone. Note, however, that using semantic features only does not perform so well on the other two test datasets, especially on the last one.

The linguistic features work relatively well on two of the test datasets, but not on bTV, where the combination of ``Credibility + Semantic'' is the best-performing one.

Naturally, the best results are on the Dnevnik vs. NE!Novinite, where the classifier achieves near perfect accuracy (note that this is despite the different class distribution on training vs. testing). The hardest testing dataset is bTV, where both the positive and the negative class are from sources different from those used in the training dataset; yet, we achieve up to 75.91\% accuracy, which is well above the majority class baseline of 50.36. The Dnevnik vs. Bazikileaks dataset falls somewhere in between, with up to 85.53\% accuracy; this is to be expected as the positive examples come from the same source as for the training dataset (even though the negative class is different). 

Overall, on all three datasets, we achieved accuracy of 75-99\%, which is well above the majority class baseline. The strong relative performance on the three different test datasets that come from different sources suggests that our model really learns to distinguish credible vs. fake news rather than learning to classify topics, sources, or author style.

\section{Conclusion and Future Work}
\label{sec:futurework}

We have presented a feature-rich language-independent approach for distinguishing credible from fake news. 
In particular, we used linguistic ($n$-gram), credibility-related (capitalization, punctuation, pronoun use, sentiment polarity, etc., with feature selection), and semantic (embeddings and DBPedia data) features.
Our experiments on three different testsets, derived from four different sources, have shown that our model can distinguish credible from fake news with very high accuracy, well above a majority-class baseline.

In future work, we plan to experiment with more features, e.g., based on linked data \cite{Zaharia:2010:SCC:1863103.1863113}, or on discourse analysis \cite{Zaharia:2010:SCC:1863103.1863113}.
Looking at features used for related tasks such as humor- \cite{yang-EtAl:2015:EMNLP2} and rumor-related \cite{zubiaga2015analysing} is another promising direction for future work.
We also want to apply deep learning, which can eliminate the need for feature engineering altogether.

Last but not least, we would like to note that we have made our source code and datasets publicly available for research purposes at the following URL:

\begin{center}
\url{https://github.com/mhardalov/news-credibility}
\end{center}

\newpage

\paragraph{Acknowledgments.}
This research was performed by Momchil Hardalov, a student in Computer Science in the Sofia University ``St Kliment Ohridski'', as part of his MSc thesis.
It is also part of the Interactive sYstems for Answer Search (Iyas) project, which is developed by the Arabic Language Technologies (ALT) group at the Qatar Computing Research Institute (QCRI), HBKU, part of Qatar Foundation in collaboration with MIT-CSAIL.

\bibliographystyle{plain}
\bibliography{bibliography}

\end{document}